\documentclass[journal]{IEEEtran}

\ifCLASSINFOpdf

\else

\fi
\usepackage{amssymb}
\usepackage{graphicx}
\usepackage{multirow}
\graphicspath{{figures/}}
\usepackage{color}
\usepackage{amsmath,amsfonts}
\usepackage{algorithmic}
\usepackage{algorithm}
\usepackage{array}
\usepackage[caption=false,font=normalsize,labelfont=sf,textfont=sf]{subfig}
\usepackage{textcomp}
\usepackage{stfloats}
\usepackage{url}
\usepackage{verbatim}
\usepackage{graphicx}
\usepackage{cite}
\usepackage{fancyhdr}

\begin{document}

\title{IVGF: The Fusion-Guided Infrared and Visible General Framework}

\author{Fangcen Liu,
        Chenqiang Gao{*} \thanks{{*}Corresponding author: Chenqiang Gao.},
        Fang Chen,
        Pengcheng Li,
        Junjie Guo,
        Deyu Meng
        
\thanks{Fangcen Liu, Pengcheng Li, and Junjie Guo are with the School of Communications and Information Engineering, Chongqing University of Posts and Telecommunications, Chongqing 400065, China (E-mail: liufc67@gmail.com, lipengchengme@163.com, s220101036@stu.cqupt.edu.cn).}

\thanks{Fang Chen is with the Department of Electrical Engineering and Computer Science, School of Engineering, University of California, Merced, 5200 Lake Rd, Merced, CA 95343, USA. E-mail: fchen20@ucmerced.edu.}

\thanks{Chenqiang Gao is with the School of Intelligent Systems Engineering, Sun Yat-sen University, Shenzhen, Guangdong 518107, China. E-mail: gaochq6@mail.sysu.edu.cn.}
 
\thanks{Deyu Meng is with the School of Mathematics and Statistics, Xi’an Jiaotong University, Xi’an, Shanxi, 710049, China. (E-mail: dymeng@mail.xjtu.edu.cn).}
}

\markboth{}%
{Shell \MakeLowercase{\textit{et al.}}: A Sample Article Using IEEEtran.cls for IEEE Journals}

\IEEEpubid{}

\maketitle
\begin{abstract}
Infrared and visible dual-modality tasks such as semantic segmentation and object detection can achieve robust performance even in extreme scenes by fusing complementary information.
Most current methods design task-specific frameworks, which are limited in generalization across multiple tasks.
In this paper, we propose a fusion-guided infrared and visible general framework, IVGF, which can be easily extended to many high-level vision tasks.
Firstly, we adopt the SOTA infrared and visible foundation models to extract the general representations.
Then, to enrich the semantics information of these general representations for high-level vision tasks, we design the feature enhancement module and token enhancement module for feature maps and tokens, respectively.
Besides, the attention-guided fusion module is proposed for effectively fusing by exploring the complementary information of two modalities.
Moreover, we also adopt the cutout\&mix augmentation strategy to conduct the data augmentation, which further improves the ability of the model to mine the regional complementary between the two modalities.
Extensive experiments show that the IVGF outperforms state-of-the-art dual-modality methods in the semantic segmentation and object detection tasks.
The detailed ablation studies demonstrate the effectiveness of each module, and another experiment explores the anti-missing modality ability of the proposed method in the dual-modality semantic segmentation task.

\end{abstract}


\IEEEpeerreviewmaketitle

\section{Introduction}
Infrared and visible imaging are two common modalities in visual perception. 
Infrared images utilize the thermal radiation of objects for imaging, effectively mitigating the adverse effects of low-illumination environments, but they lack color and detailed information \cite{liu2024infmae, MFPT }.
In contrast, visible images are formed by the reflection of objects, making them rich in color, texture, and details.
However, they are less effective in adverse environments \cite{lv2024context, zhou2023wavenet, tan2021night}.
The distinct imaging characteristics of these modalities make them highly complementary, enabling dual-modality imaging to achieve all-weather perception capabilities \cite{zhao2023mitigating, Lrrnet, IBFusion}.

\begin{figure}
    \centering
    \includegraphics[width=1.0\columnwidth]{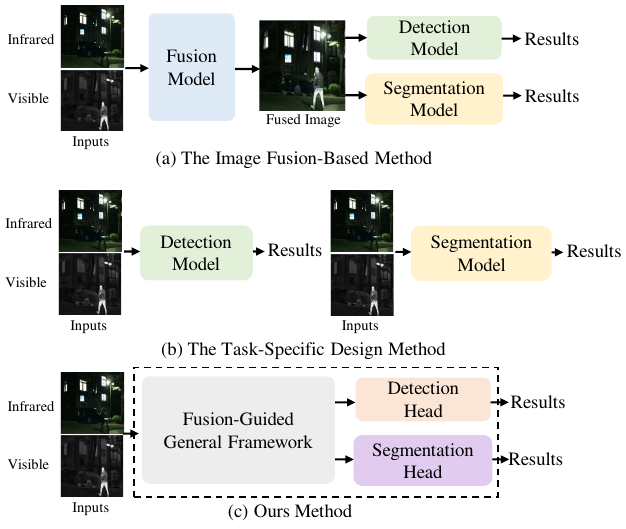}
    \caption{Different structures for infrared and visible dual-modality high-level vision tasks include:
    (a) The image fusion-based methods usually obtain the fused images and then take them as inputs for downstream tasks.
    The main purpose of these methods is to obtain high-quality fusion images.
    (b) The task-specific design methods aim to address specific problems in a single task and propose novel frameworks.
    The generalization of these methods is limited.
    (c) The proposed general framework for infrared and visible modalities exhibits great generalization, easily being extended to both detection and semantic segmentation tasks by integrating with task heads.
    }
    \label{fig:introduction}
\end{figure}

Nowadays, research on all-weather high-level vision tasks in infrared and visible modalities can be primarily divided into two approaches: the image fusion-based method and the task-specific design method, as shown in Fig. \ref{fig:introduction}.
The image fusion-based methods typically design an image fusion framework and then take the fused images as the input of the existing detection/segmentation methods \cite{HitFusion, Tang2023PSFusion}.
Moreover, some works utilize the high-level vision task to guide fusion methods \cite{liu2022target, zhao2023metafusion}.
However, the main purpose of these methods is to obtain high-quality fused images.
The task-specific design methods aim to fully utilize the complementary information of dual modalities to improve the performance of high-level vision tasks \cite{GAFF, SMPD}.
These methods can achieve results beyond the reach of a single modality, thereby enhancing the model's ability to analyze and understand the environment \cite{fan2024querytrack, liang2024multi, li2024m2fnet, fu2023lraf, zhang2023differential, li2023sosmaskfuse}.
Although these task-specific designs can obtain superior performance on a single task, they are challenging to be generalized across multiple tasks.

To address the challenge of limited generalization, the foundation model has been proposed.
These foundation models \cite{EVA2023, mae2022, liu2024infmae, jin2024efficient} demonstrate powerful learning capabilities of general representations and can achieve results surpassing task-specific design methods by integrating with existing task heads \cite{Wu_2024_CVPR}.
Such foundation models can improve the performance of various high-level vision tasks and reduce the design complexity \cite{SpectralGPT}. 
Currently, there are relevant foundation models in the single modality, such as infrared \cite{liu2024infmae}, remote sensing \cite{satmae2022}, and video \cite{wang2023videomae}.
Besides, many foundation models have also emerged in multimodal communities such as text-image and video-image \cite{mplug, CREPE}. 
However, research within the infrared and visible dual-modal community remains limited, and training such models requires millions of well-aligned images.
Obtaining these images is challenging because the dual-modal images are usually captured asynchronously, and the fields of view of dual-modal sensors differ significantly.
Moreover, aligning these images accurately is extremely labor-intensive and time-consuming.
Hence, it is natural to come out with a question: can we build an infrared and visible general framework upon the existing foundation models that can be seamlessly integrated with high-level vision tasks without a large number of paired images?

Intuitively, directly summing the features from an infrared foundation model with those from a visible foundation model appears to be a viable solution.
However, as shown in Table \ref{tab:ablation} (baseline), our experimental findings suggest this approach is suboptimal.
It may be because simply combining the general features of infrared and visible foundation models remains two issues.
(1) The general features need to enrich the semantic information further.
(2) Summing the features can not fully integrate the complementary information between the two modalities.

To this end, we explore the characteristics of infrared and visible images and propose a fusion-guided Infrared and Visible General Framework (IVGF) by utilizing the state-of-the-art foundation models.
Firstly, to avoid the requirement for a large number of well-aligned images, we directly combine our infrared foundation model InfMAE \cite{liu2024infmae} and the well-designed visible foundation model MCMAE \cite{MCMAE2022} as our basic framework.
Then, to explore more semantic information, we design the feature enhancement module and the token enhancement module.
The feature enhancement module is designed to enhance the semantic information of the feature maps obtained from the encoder blocks of the single-modality foundation models, while the token enhancement module is suitable for the tokens obtained in the encoder block3.
Besides, we design an attention-guided fusion module that adopts the cross-modality attention mechanism to fuse visible and infrared features based on their complementary characteristics.
Moreover, to help further mine the regional complementary between the two modalities, we design the cutout\&mix augmentation, which encourages the model to recover the masked information with another modality and exposes the model to more diverse samples during training.

We summarize the main contributions of the paper as follows:
\begin{itemize}
     \item We propose a simple and efficient fusion-guided infrared and visible general framework (IVGF), which does not require a large amount of well-aligned images and effectively improves the generalization for dual-modality downstream tasks.
     \item We design the feature enhancement module and the token enhancement module, which are suited for feature maps and tokenized sequences, respectively, to enhance their semantic information. 
     Besides, we fully explore and fuse the complementary characteristics of infrared and visible modalities by designing an attention-guided fusion module.
     \item The experimental results show that the method proposed in this paper outperforms state-of-the-art methods, achieving the best performance in downstream tasks.
 \end{itemize}

The remainder of this paper is organized as follows:
In Section \ref{sec:related work}, related works are briefly reviewed.
In Section \ref{sec:method}, we present the proposed method in detail.
In Section \ref{sec:experiment}, the experimental results are given and discussed.
Conclusions are drawn in Section \ref{sec:conclude}.

\section{Related Work}
\label{sec:related work}

\subsection{The Foundation Model}
The foundation model can be classified as single-modality and multi-modality foundations.

\subsubsection{Single-Modality Foundation Models}
ViT \cite{VIT} as a breakthrough work in the computer vision community has laid the foundation for developing foundation models.
In the field of visible modality, models such as EVA \cite{EVA2023}, MAE \cite{mae2022}, and MCMAE \cite{MCMAE2022} have achieved outstanding performance in various tasks such as object detection, segmentation, and classification.
In the field of remote sensing modality, models such as SatMAE \cite{satmae2022} and MTP \cite{wang2024mtp} have reduced the design complexity of tasks like remote sensing object classification, segmentation, and recognition.
Recently, in the field of infrared modality, the foundation model InfMAE \cite{liu2024infmae} has been introduced, achieving state-of-the-art performance in tasks such as semantic segmentation, object detection, and small object detection.
In addition to multi-task foundation models, models like SAM \cite{SAM2023}, MedSAM \cite{ma2024MedSAM}, and Rein \cite{Rein} are specifically designed for the image segmentation task.

\subsubsection{Multi-Modality Foundation Models}
In recent years, benefiting from the efficient scalability and modality inclusiveness of the Transformer structure, a large number of multi-modality foundational models have emerged.
As a pioneer in multimodal foundational models, CLIP \cite{Clip2021} has made significant contributions to various tasks in the language-visual domain.
Subsequently, many multimodal foundational models have been built upon the CLIP.
For instance, Meta-Transformer \cite{zhang2023metatransformer} and Onellm \cite{han2023onellm} leveraged the pre-trained weights of CLIP's text encoder to establish a multi-modality, multi-task integration framework. 
Moreover, MultiMAE \cite{bachmann2022multimae} was inspired by the idea of masking image modeling, and accomplished multi-modality, multi-task learning through masked self-supervised learning.
Additionally, this approach supports both single-modality and multi-modality inputs concurrently.

The aforementioned single-modality and multi-modality methods require a large number of images for self-supervised learning.
However, due to the lack of well-aligned infrared and visible images, the exploration of foundation models in the infrared and visible dual-modal community remains in its infancy.
In this paper, we propose a first infrared and visible general framework based on existing foundation models, fully leveraging the complementarity of the two modalities and avoiding the requirement for hundreds of paired images.

\subsection{The Infrared and Visible Tasks}
The approaches to infrared and visible tasks can be roughly divided into the image fusion-based method and the task-specific design method.
\subsubsection{The Image Fusion-Based Methods}
These methods maintain the merits of infrared and visible modalities to obtain high-quality fused images, thereby improving the performance of downstream semantic understanding tasks.

Some of them fuse the images first and take them as the inputs of the existing segmentation and detection methods \cite{2024dispel, zhao2023cddfuse}.
Specifically, Zhang et al. \cite{2024dispel} proposed a controllable visual enhancer to dispel darkness and obtain images with high visual quality.
These images are input into existing object detection and semantic segmentation methods and show their effectiveness.
Zhao et al. \cite{zhao2023cddfuse} proposed a correlation-driven feature decomposition fusion method to maintain the advantage of different modalities.

Some other methods combine the fusion task with object detection or segmentation tasks by employing joint learning \cite{liu2022target, HitFusion, zhao2023metafusion, Tang2023PSFusion, li2023sosmaskfuse}.
Liu et al. \cite{liu2022target} proposed an object-aware fusion and detection network, which achieved robust object detection and fusion through bidirectional adversarial learning.
Chen et al. \cite{HitFusion} designed a three-stage training strategy to make the fusion framework more suitable for high-level vision tasks.
Zhao et al. \cite{zhao2023metafusion} bridged the feature gap between low-level fusion and high-level detection tasks by designing the meta-feature embedding.
Tang et al. \cite{Tang2023PSFusion} proposed a semantic-driven fusion model based on progressive semantic injection and scene fidelity constraints, improving high-level vision task performance.
Li et al. \cite{li2023sosmaskfuse} proposed a two-stage network, SOSMaskFuse, which effectively integrates important information from bimodal sources and performs detection tasks.

\subsubsection{The Task-Specific Design Methods}
The infrared-visible semantic segmentation and object detection are the most common tasks.
We briefly review some outstanding works.

\textbf{The Infrared and Visible Image Semantic Segmentation Task}:
Infrared and visible semantic segmentation ensures reliable performance under extreme conditions such as overexposure, smoke, and darkness \cite{liu2023multi}. 
Currently, the mainstream approach involves designing effective fusion modules that leverage the complementary and differential characteristics of dual-modal images.
Lv et al. \cite{lv2024context} explored the complementary relationship of dual-modal images by designing a context-aware interaction network, achieving promising results on public infrared and visible semantic segmentation datasets.
Zhang et al. \cite{zhang2021ABMDRNet} adopted a bridging-then-fusing strategy to achieve multi-level fusion of dual-modal features, reducing the disparity between the two modalities.
Shin et al. \cite{crm} proposed a complementary random masking strategy and the self-distillation loss to prevent over-reliance on a single modality and to extract complementary and meaningful representations, thereby enhancing the accuracy and robustness of the network.
Zhang et al. \cite{CMX} designed a unified fusion framework for RGB-X semantic segmentation tasks.
Specifically, the cross-modal feature rectification module is designed to calibrate dual-modal features, and the feature fusion module is deployed to facilitate a sufficient exchange of long-range contexts before mixing.
Zhang et al. \cite{zhang2023delivering} extensively explored semantic segmentation models integrating an arbitrary number of modalities, yielding good segmentation results under various adverse conditions.

\textbf{The Infrared and Visible Image Object Detection Task}:
Fu et al. \cite{fu2023lraf} considered the potential interaction of long-range dependence between different modalities and proposed a feature-enhanced long-range attention fusion network (LRAF-Net), which improves detection performance by integrating the long-range dependencies of enhanced visible and infrared features.
Cao et al. \cite{cao2023multimodal} proposed a lightweight fusion module capable of fusing inputs from different modalities through the designed channel switching and spatial attention mechanisms.
Guo et al. \cite{guo2024damsdet} considered the interference between modalities and designed MS-DETR, which can adaptively select high-confidence targets in environmental interference, further enhancing the robustness of object detection.
Zhang et al. \cite{zhang2023saliency} proposed a bi-modal salient detection architecture called saliency prototype network (SPNet), which dynamically allocates fusion weights during the fusion process.

Different from the above methods, the proposed method is a general framework that can be extended to both semantic segmentation and object detection tasks by integrating with the task-specific head.

\section{Method}
\label{sec:method} 
\subsection{Overview}
\begin{figure*}
    \centering
    \includegraphics[width=1.0\textwidth]{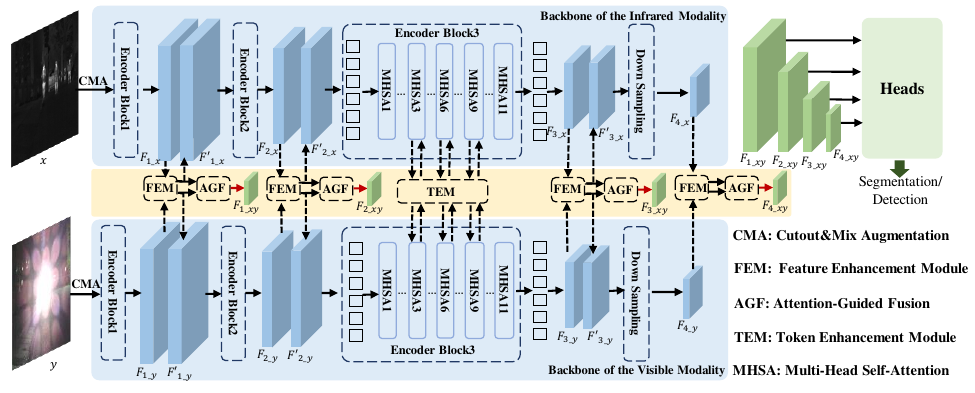}
    \caption{The framework of the proposed method. 
    It contains five main components: two modality-specific backbones, the feature enhancement module, the token enhancement module, the attention-guided fusion module, and the task-specific head.
    The backbone can extract the general representation of each modality.
    The feature enhancement and token enhancement modules are designed for feature maps and tokens, respectively.
    The attention-guided fusion module integrates the complementary information from two modal features.
    }
    \label{fig:pipline}
\end{figure*}

The overview of our method is illustrated in Fig. \ref{fig:pipline}.
Our method contains five main components: two modality-specific backbones, the feature enhancement module (FEM), the attention-guided fusion (AGF) module, the token enhancement module (TEM), and task-specific heads.
Moreover, we design a cutout\&mix augmentation (CMA), which helps to mine the regional complementary between the two modalities.
Specifically, we choose the well-trained encoders of InfMAE \cite{liu2024infmae} and MCMAE \cite{MCMAE2022} as backbones for infrared and visible modalities, respectively.
Firstly, the infrared image $x$ and visible image $y$ conduct data augmentation by the proposed cutout\&mix augmentation.
After that, we feed them into the backbones to obtain the general representations $F_{1_{-}x}$, $F_{2_{-}x}$, $F_{3_{-}x}$, $F_{4_{-}x}$, and $F_{1_{-}y}$, $F_{2_{-}y}$, $F_{3_{-}y}$, $F_{4_{-}y}$.
During this process, we propose the feature enhancement to enrich the semantic representations for the general representations.
At the same time, we design the token enhancement module to enrich the semantic information of the tokens, which are obtained in the encoder block3.
After that, the attention-guided fusion module is proposed to fuse the dual-modality features according to their complementary characteristics and obtain the fused features $F_{1_{-}xy}$, $F_{2_{-}xy}$, $F_{3_{-}xy}$ and $F_{4_{-}xy}$.
Finally, the fused features are input into the task-specific head to get the segmentation/detection results.

\subsection{The Feature Enhancement Module}
\label{FEM}

\begin{figure}
    \centering
    \includegraphics[width=0.5\textwidth]{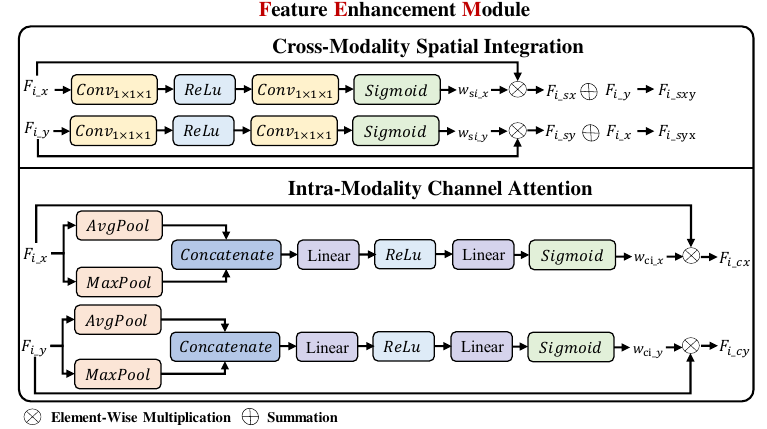}
    \caption{The proposed feature enhancement module.
    It contains the cross-modality spatial integration and the intra-modality channel attention.
    }
    \label{fig:fem}
\end{figure}

Due to the different imaging characteristics of infrared and visible images, the designed feature enhancement module enriches the semantic information of feature maps from different dimensions.
Fig. \ref{fig:fem} shows the specific module design.

Initially, different image regions can provide different contributions to the task, and such differences are also present between the two modalities.
Hence, we enhance the two model features through cross-modality spatial integration. 
Specifically, we feed the features $F_{i_{-}x} \in \mathbb{R}^{C_{i} \times H_{i} \times W_{i}}$ and $F_{i_{-}y} \in \mathbb{R}^{C_{i} \times H_{i} \times W_{i}}$ which are obtained from the backbones into the cross-modality spatial attention to obtain the spatial attention weights $W_{si_{-}x} \in \mathbb{R}^{1 \times H_{i} \times W_{i}}$, $W_{si_{-}y} \in \mathbb{R}^{1 \times H_{i} \times W_{i}}$.
We can formulate it as follows,
\begin{equation}
\scriptsize
    W_{si_{-}x} = Sigmoid(Conv(ReLu(Conv(F_{i_{-}x})))),
\end{equation}   
\begin{equation}
\scriptsize
     W_{si_{-}y} = Sigmoid(Conv(ReLu(Conv(F_{i_{-}y})))),
\end{equation}  
where the `$i$' means the $i-th$ encoder block in the backbone.
The $Conv(\cdot)$ represents the convolution with the kernel size of 1$\times$1, the stride of 1.
The $ReLu(\cdot)$ and the $Sigmoid(\cdot)$ are activation functions.
Then, we multiply the $F_{i_{-}x}$ and $F_{i_{-}y}$ with $W_{si_{-}x}$ and $W_{si_{-}y}$ to obtain $F_{i_{-}sx} \in \mathbb{R}^{C_{i} \times H_{i} \times W_{i}} $ and $F_{i_{-}sy} \in \mathbb{R}^{C_{i} \times H_{i} \times W_{i}}$.
We can formulate it as follows,
\begin{equation}
    F_{i_{-}sx} = F_{i_{-}x} \times W_{si_{-}x},
\end{equation}   
\begin{equation}
     F_{i_{-}sy} = F_{i_{-}y} \times W_{si_{-}y},
\end{equation}  
where the `$\times$' is the element-wise multiplication.
Finally, to avoid missing information and realize cross-modality spatial integration, we add the input features $F_{i_{-}x}$ and $F_{i_{-}y}$ with $F_{i_{-}sy}$ and $F_{i_{-}sx}$, respectively.
We can formulate it as follows,
\begin{equation}
    F_{i_{-}sxy} = F_{i_{-}x} + F_{i_{-}sy},
\end{equation}   
\begin{equation}
     F_{i_{-}syx} = F_{i_{-}y} + F_{i_{-}sx}.
\end{equation} 

In the channel dimension, we adopt the intra-modality channel attention to capture the importance of each feature channel.
Taking the infrared modality as an example, the features $F_{i_{-}x}$ are firstly fed into the multi-dimensional perception module to obtain the channel-wise representations $F_{a_{-}x} \in \mathbb{R}^{C_{i} \times 1 \times 1} $, $F_{m_{-}x} \in \mathbb{R}^{C_{i} \times 1 \times 1}$.
We can formulate it as follows,
\begin{equation}
    F_{a_{-}x} = AP(F_{i_{-}x}),
\end{equation}  
\begin{equation}
    F_{m_{-}x} = MP(F_{i_{-}x}),
\end{equation} 
where the $AP(\cdot)$ and $MP(\cdot)$ are the adaptive average pooling and the adaptive max pooling, respectively. 
After that, we concatenate them along the channel dimension to obtain $F_x$.
Then the channel attention weights $W_{ci_{-}x}$ are calculated by the channel information perception module which contains two linear layers.
We can formulate it as follows,
\begin{equation}
    W_{ci_{-}x} = Sigmoid(LN(ReLu(LN(F_x)))),
\end{equation} 
where the the $LN(\cdot)$ represents the linear layer, and the $ReLu(\cdot)$ and the $Sigmoid(\cdot)$ are activation functions.
In the same way, the channel attention weights $W_{ci_{-}y}$ of visible modality are obtained.
After that, we multiply the $F_{i_{-}x}$ and $F_{i_{-}y}$ with $W_{ci_{-}x}$ and $W_{ci_{-}y}$ to obtain $F_{i_{-}cx} \in \mathbb{R}^{C_{i} \times H_{i} \times W_{i}}$ and $F_{i_{-}cy} \in \mathbb{R}^{C_{i} \times H_{i} \times W_{i}}$.
We can formulate it as follows,
\begin{equation}
    F_{i_{-}cx} = F_{i_{-}x} \times W_{ci_{-}x},
\end{equation}   
\begin{equation}
     F_{i_{-}cy} = F_{i_{-}y} \times W_{ci_{-}y},
\end{equation} 
where the `$\times$' is the element-wise multiplication.

Finally, we integrate the cross-model spatial enhanced features and the intra-model channel enhanced features to obtain the infrared and visible features $F_{i_{-}x}^{'} \in \mathbb{R}^{C_{i} \times H_{i} \times W_{i}}$ and $F_{i_{-}y}^{'} \in \mathbb{R}^{C_{i} \times H_{i} \times W_{i}}$ for the next encoder block. We can formulate it as follows,
\begin{equation}
    F_{i_{-}x}^{'}=F_{i_{-}sxy} + F_{i_{-}cx},
\end{equation}
\begin{equation}
    F_{i_{-}y}^{'}=F_{i_{-}syx} + F_{i_{-}cy}.
\end{equation}

After the feature enhancement module, we obtain the features with more powerful semantic representations.
It should be mentioned that we reshape the output tokens of the encoder block3 to conduct feature enhancement, and then obtain $F_{3\_{x}}^{'} \in \mathbb{R}^{C_3 \times H_3 \times W_3}$ and $F_{3\_{y}}^{'} \in \mathbb{R}^{C_3 \times H_3 \times W_3}$ for attention-guided fusion module.

\subsection{The Token Enhancement Module}
\label{PA}
\begin{figure}
    \centering
    \includegraphics[width=0.5\textwidth]{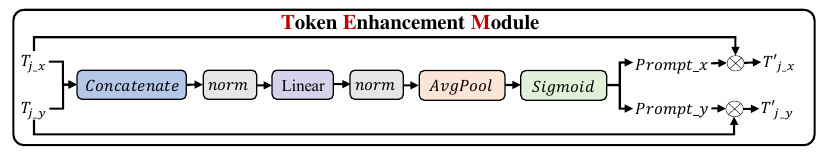}
    \caption{The proposed token enhancement module.
    }
    \label{fig:tem}
\end{figure}

Different from the feature enhancement module, the token enhancement module is designed for the tokenized features in the encoder block3.
As shown in Fig. \ref{fig:tem}, the infrared and visible features $T_{j_{-}x}$ and $T_{j_{-}y}$ are obtained for subsequent processing, where the $j$ is the $j-th$ multi-head self-attention (MHSA) layer and we set $j$ to 3, 6, and 9.

In the token enhancement module, we concatenate $T_{j_{-}x} \in \mathbb{R}^{N \times C}$ and $T_{j_{-}y} \in \mathbb{R}^{N \times C}$ along the $C$ dimension as a merged representation of dual-model tokens.
Then after a layer normalization, we adopt a linear layer to reduce feature dimension and obtain the dual-model features $\Phi \in \mathbb{R}^{N \times C_1}$.
We can formulate these processes as follows,
\begin{equation}
    \Phi=L(Norm(Cat(T_{j_{-}x}, T_{j_{-}y}))),
\end{equation}
where $Cat(\cdot)$ represents the concatenation of features along the channel dimension.
The $norm(\cdot)$ is the layer normalization operation, and $L(\cdot)$ is a linear process.
After that, we employ the mixture of adapters \cite{zhu2024taskcustomized} to estimate each token preference, it helps sense the importance of each token, and then the importance prompt of each modality is generated by the following processing,
\begin{equation}
    \{promtp_{\_x}, prompt_{\_y}\}=Sigmoid(AP(\Phi)),
\end{equation}
where $promtp_{\_x} \in \mathbb{R}^{N \times 1}$, $promtp_{\_y} \in \mathbb{R}^{N \times 1}$. 
The $AP(\cdot)$ represents the adaptive average pooling, and the $Sigmoid(\cdot)$ is an activation function.
Finally, we conduct the element-wise multiplication between importance prompts with the features $T_{j_{-}x}$ and $T_{j_{-}y}$ to obtain the attention tokens.
We can formulate it as follows,
\begin{equation}
    T_{j_{-}x}^{'} = T_{j_{-}x} \times promtp_x,
\end{equation}
\begin{equation}
    T_{j_{-}y}^{'} = T_{j_{-}y} \times promtp_y.
\end{equation}
The $T_{j_{-}x}^{'} \in \mathbb{R}^{N \times C}$ and $T_{j_{-}y}^{'} \in \mathbb{R}^{N \times C}$ express tokens with a powerful semantic information.

\subsection{The Attention-Guided Fusion Module}
\label{AGF}

\begin{figure}
    \centering
    \includegraphics[width=0.5\textwidth]{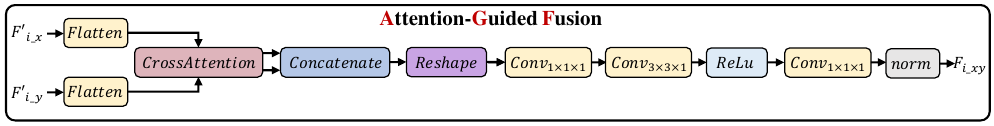}
    \caption{The proposed attention-guided fusion module.
    }
    \label{fig:agf}
\end{figure}

The attention-guided fusion module aims to explore the relationship between two modalities and extract the complementary from each other modality.
The details of the attention-guided fusion module can be seen in Fig. \ref{fig:agf}.

We feed $F_{i_{-}x}^{'} \in \mathbb{R}^{C_i \times H_i \times W_i}$ and $F_{i_{-}y}^{'} \in \mathbb{R}^{C_i \times H_i \times W_i}$ into the attention-guided fusion module to obtain the fused features $F_{i_{-}xy} \in \mathbb{R}^{C_i \times H_i \times W_i}$.
Specifically, we flatten $F_{i_{-}x}^{'}$ and $F_{i_{-}y}^{'}$, geting $R_{i_{-}x} \in \mathbb{R}^{C_i \times HW_i}$ and $R_{i_{-}y} \in \mathbb{R}^{C_i \times HW_i}$.
Then we adopt the cross-attention to extract the complementary information from another modality.
We can formulate it as follows,
\begin{equation}
    CrossAttn_{xy}\left(Q_{x}, K_{y}, V_{y}\right)=softmax\left(\frac{Q_{x} K_{y}{ }^{T}}{\sqrt{d_{k}}}\right) V_{y},
\end{equation}
\begin{equation}
    CrossAttn_{yx}\left(Q_{y}, K_{x}, V_{x}\right)=softmax\left(\frac{Q_{y} K_{x}{ }^{T}}{\sqrt{d_{k}}}\right) V_{x},
\end{equation}
where the $Q, K, V$ are the quarry, key, and value which are obtained by the linear layers.
After that we fusion these features by the merge module $M(\cdot)$, the specific operations are as follows,
\begin{equation}
    F_{i_{-}xy}=M(Re(Cat(CrossAtt_{xy}, CrossAtt_{yx}))),
\end{equation}
where $Cat(\cdot)$ is the concatenation of features along the channel dimension, the $Re(\cdot)$ means that we reshape the tokens to the shape of $C_i \times H_i \times W_i$.
The $M(\cdot)$ consists of two convolution layers with both the kernel size of 1$\times$1, a convolution layer with a kernel size of 3$\times$3, and a ReLu activation function.

The fused features $F_{1_{-}xy}$, $F_{2_{-}xy}$, $F_{3_{-}xy}$ and $F_{4_{-}xy}$ contain the complementary information of two modalities and have a more powerful representation ability for downstream tasks.

\subsection{The Cutout\&mix Augmentation}
\label{DA}
\begin{figure}[h]
    \centering
    \includegraphics[width=0.5\textwidth]{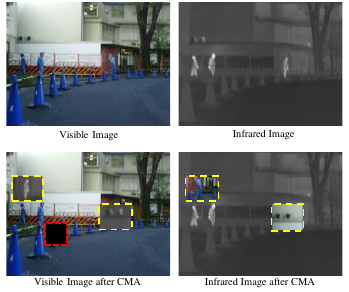}
    \caption{The proposed cutout\&mix augmentation.
    The yellow gridlines indicate the cutmix augmentation and the rad gridline is the random cutout augmentation.
    }
    \label{fig:CMA}
\end{figure}

The key for infrared and visible tasks is to mine the regional complementary between the two modalities.
Hence, we design the cutout\&mix augmentation (CMA) to encourage the model to recover the masked information with another modality and allow it to see more diverse samples during training, thereby improving its generalization ability.
This strategy consists of two parts: cutout augmentation and cutmix augmentation.
The cutout augmentation is designed to facilitate the learning of adaptive cross-modality regional complementarity, the artificial optically-impaired regions are randomly created in a random modality. 
The cutmix augmentation involves randomly cutting and pasting patches between two model images, thus creating new training samples.
An example is shown in Fig. \ref{fig:CMA}.

\section{Experiment}
\label{sec:experiment}

\subsection{Datasets and Evaluation Metrics}
We conduct experiments on two downstream tasks: semantic segmentation and object detection.

\textbf{Semantic segmentation:}
MFNet \cite{ha2017mfnet} is the visible-infrared urban scene image dataset with pixel-level annotation. 
This dataset contains 1569 images (820 taken at daytime and 749 taken at nighttime). 
Eight classes of objects commonly encountered during driving are labeled in this dataset.

MSRS \cite{Tang2022PIAFusion} provides 1444 paired infrared and visible images. 
The training set contains 1083 pairs of infrared and visible images and the testing set consists of 361 image paires.
This dataset also contains night classes.
The differences between the MFNet and MSRS datasets are that the MSRS dataset removed 125 pairs of clearly misaligned images from the MFNet dataset and applied contrast stretching to the infrared images.

We adopt the mean Intersection over Union (mIoU) of all classes for evaluation.

\textbf{Object Detection:}
M3FD \cite{liu2022target} is an infrared and visible target detection dataset and contains 4200 images. 
We split it into a training set of 3,780 pairs and a test set of 420 pairs. 
The dataset labels six targets during driving.

We evaluate the object detection performance with $mAP_{50-95}$ and $mAP_{50}$ of all classes.

\begin{table*}[]
\centering
\caption{The semantic segmentation results of all the methods on the MSRS dataset.}
\label{tab:MSRS}
\resizebox{1.8\columnwidth}{!}{
\begin{tabular}{c|c|cccccccc|cc}
\hline
Methods                                             & Baseline & \begin{tabular}[c]{@{}c@{}}CDDFuse\\ \cite{zhao2023cddfuse}\end{tabular} & \begin{tabular}[c]{@{}c@{}}SpiderMesh\\ \cite{spidermesh}\end{tabular} & \begin{tabular}[c]{@{}c@{}}RSFNet\\ \cite{rsfnet}\end{tabular} & \begin{tabular}[c]{@{}c@{}}CEKD\\ \cite{CEKD}\end{tabular} & \begin{tabular}[c]{@{}c@{}}EAEFNet\\ \cite{EAEFNet}\end{tabular} & \begin{tabular}[c]{@{}c@{}}IGFNet\\ \cite{IGFNet}\end{tabular} & \begin{tabular}[c]{@{}c@{}}HAPNet\\ \cite{HAPNet}\end{tabular} & \begin{tabular}[c]{@{}c@{}}CRM\\ \cite{crm}\end{tabular} & \multicolumn{1}{c}{\begin{tabular}[c]{@{}c@{}}IVGF \\ (Scratch)\end{tabular}} & \begin{tabular}[c]{@{}c@{}}IVGF \\ (Ours)\end{tabular} \\ \hline
\begin{tabular}[c]{@{}c@{}}mIou\\ (\%)\end{tabular} & 78.6  & 64.3  & 75.2 & 73.6  & 57.4      & 75.5          & 73.5   & 70.1  & 69.8        & 72.5         & \textbf{81.0}           \\ \hline
\end{tabular}%
}
\end{table*}

\begin{table*}[]
\centering
\caption{The semantic segmentation results of all the methods on the MFNet dataset.}
\label{tab:MFNet}
\resizebox{1.8\columnwidth}{!}{
\begin{tabular}{c|c|cccccccc|cc}
\hline
Methods                                             & Baseline & \begin{tabular}[c]{@{}c@{}}CDDFuse\\ \cite{zhao2023cddfuse}\end{tabular} & \begin{tabular}[c]{@{}c@{}}SpiderMesh\\ \cite{spidermesh}\end{tabular} & \begin{tabular}[c]{@{}c@{}}RSFNet\\ \cite{rsfnet}\end{tabular} & \begin{tabular}[c]{@{}c@{}}CEKD\\ \cite{CEKD}\end{tabular} & \begin{tabular}[c]{@{}c@{}}EAEFNet\\ \cite{EAEFNet}\end{tabular} & \begin{tabular}[c]{@{}c@{}}IGFNet\\ \cite{IGFNet}\end{tabular} & \begin{tabular}[c]{@{}c@{}}HAPNet\\ \cite{HAPNet}\end{tabular} & \begin{tabular}[c]{@{}c@{}}CRM\\ \cite{crm}\end{tabular} & \multicolumn{1}{c}{\begin{tabular}[c]{@{}c@{}}IVGF \\ (Scratch)\end{tabular}} & \begin{tabular}[c]{@{}c@{}}IVGF \\ (Ours)\end{tabular} \\ \hline
\begin{tabular}[c]{@{}c@{}}mIou\\ (\%)\end{tabular} & 58.0     & 44.5                                                                      & 58.4                                                                    & 54.5                                                            & 53.7                                                        & 55.4                                                              & 54.6                                                            & 61.5                                                            & 61.4                                                                 & 44.9                                                                           & \textbf{62.4}                                                   \\ \hline
\end{tabular}%
}
\end{table*}

\subsection{Implementation Details.}
\subsubsection{Semantic Segmentation}
We take the InfMAE \cite{liu2024infmae} encoder pretrained on Inf30 as the backbone of the infrared modality, while the MCMAE \cite{MCMAE2022} encoder pretrained on Imagnet1K as the backbone of the visible modality.
We conduct the semantic segmentation task in the MMsegmentation Toolbox \cite{mmseg2020}.
We adopt the Upernet \cite{upernet} as the hierarchical segmentation network head.
The input images are set to 512$\times$512. 
The initial learning rate is 0.0001, the weight decay is 0.05, and the AdamW optimizer is used.
Following InfMAE \cite{liu2024infmae}, we use a 240k iteration polynomial learning rate scheme with the first 1500 iterations warmed up and the batch size is set to 2.

\subsubsection{Object Detection}
We take the InfMAE \cite{liu2024infmae} encoder pretrained on Inf30 as the backbone of the infrared modality, while the MCMAE \cite{MCMAE2022} encoder pretrained on Imagnet1K as the backbone of the visible modality.
We conduct the detection task in the MMdetection Toolbox \cite{mmdetection}.
We employ the MaskRCNN \cite{maskrcnn} as the detection head.
The input images are set to 896$\times$896.
The initial learning rate is 0.0001, the weight decay is 0.1, and the AdamW optimizer is used.
We use a 160k iteration polynomial learning rate scheme and set the batch size to 2.

We implement the semantic segmentation and object detection tasks using the pytorch 1.8.0 and 1.9.0, respectively, and train them on one NVIDIA RTX A100 GPU.

\subsection{IVGF on Semantic Segmentation}
\subsubsection{The Comparison Methods}
We compare the proposed method with the state-of-the-art methods, \textit{i.e.}, CDDFuse \cite{zhao2023cddfuse}, SpiderMesh \cite{spidermesh}, RSFNet \cite{rsfnet}, CEKD \cite{CEKD}, EAEFNet \cite{EAEFNet}, IGFNet \cite{IGFNet}, HAPNet \cite{HAPNet}, CRM \cite{crm}.
The CDDFuse is an image fusion-based method and the rest are task-specific design methods.

\subsubsection{Results on MSRS}
All the semantic segmentation results are reported in Table \ref{tab:MSRS}.
We can see from the table that compared with the CDDFuse, the proposed method outperforms it a lot.
Besides, our method performs much better than other task-specific design methods.
This may be because our method could mine more comprehensive complementary information during the end-to-end training.
Specifically, the proposed method outperforms EAEFNet \cite{EAEFNet} by 5.5\%.
Compared with other methods, our method improves more than 5.5\% on the metric of mIoU.
This significant performance improvement shows that our method helps to extract powerful semantic representations.

Meanwhile, we also train the semantic segmentation task from scratch by using our framework.
The result shows that well-trained backbones are important for the downstream task. 

\begin{table*}[]
\centering
\caption{The object detection results of all the methods on the M3FD dataset.}
\label{tab:M3FD}
\resizebox{1.4\columnwidth}{!}{
\begin{tabular}{c|c|ccccc|cc}
\hline
Methods                                                      & Baseline & \begin{tabular}[c]{@{}c@{}}CDDFuse\\  \cite{zhao2023cddfuse}\end{tabular} & \begin{tabular}[c]{@{}c@{}}ICAFusion \\ \cite{ICAFusion}\end{tabular} & \begin{tabular}[c]{@{}c@{}}CBAM \\ \cite{CBAM}\end{tabular} & \begin{tabular}[c]{@{}c@{}}CFT \\ \cite{CFT}\end{tabular} & \begin{tabular}[c]{@{}c@{}}CoCoNet\\  \cite{liu2024coconet}\end{tabular} & \begin{tabular}[c]{@{}c@{}}IVGF \\ (Scratch)\end{tabular} & \begin{tabular}[c]{@{}c@{}}IVGF \\ (Ours)\end{tabular} \\ \hline
\begin{tabular}[c]{@{}c@{}}$mAP50$\\ (\%)\end{tabular}          & 81.4     & 81.0                                                                       & 67.8                                                                   & 81.0                                                         & 68.2                                                       & 80.7                                                                      & 73.9                                                                           & \textbf{82.8}                                                   \\
\begin{tabular}[c]{@{}c@{}}$mAP_{50-95}$\\ (\%)\end{tabular} & 52.3     & 52.0                                                                       & 41.9                                                                   & 50.5                                                         & 42.5                                                       & 54.2                                                                      & 45.0                                                                           & \textbf{55.8}                                                   \\
 \hline
\end{tabular}%
}
\end{table*}

\subsubsection{Results on MFNet}
This dataset is challenging due to the low illumination scenes and the poor imaging quality of the infrared images.
The experimental results are shown in Table \ref{tab:MFNet}.
We find that the proposed method outperforms other SOTA methods.
Specifically, the proposed method outperforms HAPNet on mIoU by 0.9\%, and compared with others, our method improves more than 0.9\% on the metric of mIoU.
These results demonstrate that our method can achieve better performance on this challenging dataset.

\subsection{IVGF on Object Detection}
We compared the proposed method with the state-of-the-art methods, \textit{i.e.}, CDDFuse \cite{zhao2023cddfuse}, ICAFusion \cite{ICAFusion}, CBAM \cite{CBAM}, CFT \cite{CFT}, and CoCoNet \cite{liu2024coconet}.

We evaluate the above methods on the M3FD dataset, and the experimental results are reported in Table \ref{tab:M3FD}.
This dataset is very challenging with different scenes.
Hence, all the methods have a limited performance.
However, we can see that our method outperforms other dual-modality methods.
Specifically, our method outperforms CFT\cite{CFT} by 14.6\% on $mAP_{50}$.
It also outperforms CoCoNet \cite{liu2024coconet} by 2.1\% on $mAP_{50}$.
These results show that our method can better adapt to complex and changeable scenes.

\subsection{Ablation Study of IVGF}
To show the effectiveness of key modules and the proposed data augmentation strategy of our method, we perform ablation experiments on semantic segmentation and object detection tasks with MFNet and M3FD datasets, respectively.
We define the feature enhancement module as FEM, the attention-guided fusion module as AGF, the token enhancement module as the TEM, and the cutout\&mix augmentation as the CMA.
The experimental results can be shown in Table \ref{tab:ablation}.
The baseline method shown in line 1 is that we only adopt the backbone encoders of InfMAE and MCMAE to extract the multi-scale features and fuse them by summing.
We analyze the effectiveness of each module and strategy in the following sections.

\begin{table}[]
\centering
\caption{The effectiveness of key modules and the proposed strategy of the proposed method.
The `FEM', `TEM', `AGF', and `CMA' are the feature enhancement module, token enhancement module, attention-guided fusion, and cutout\&mix augmentation strategy, respectively.}
\label{tab:ablation}
\resizebox{1\columnwidth}{!}{
\begin{tabular}{ccccccc}
\hline
 Lines  &FEM  & AGF & TEM & CMA & Seg\_mIoU & Det\_$mAP_{50-95}$ \\ \hline
1&-            & -   & -   & -  &  58.0         & 52.3         \\  
2           & \checkmark   & - &-    & -   & 59.9          & 53.6         \\   
3            & -    & \checkmark   & - &-  & 59.6          &  53.8        \\ 
4           & -    & -    & \checkmark &-  & 59.6          &  53.6        \\
5 &-    & -   & -    & \checkmark   & 59.7          & 53.7         \\
6           & \checkmark   & \checkmark  &- & -   &  60.1         & 55.4         \\  
7            & -    & \checkmark   & \checkmark &-  &  60.4         & 55.1         \\
8          & \checkmark   & -    & \checkmark &-  &   60.2        & 54.4         \\  
9   & \checkmark   & \checkmark   & \checkmark &-      &  61.1  & 55.5             \\ \hline
10 &\checkmark            & \checkmark   & \checkmark   & \checkmark  & 62.4          &  55.8        \\ \hline
\end{tabular}
}
\end{table}

\subsubsection{Effect of the Feature Enhancement Module (FEM)}
When we only adopt the FEM (line 2), we fuse the features by summing.
At this time, as shown in Table \ref{tab:ablation},  the mIoU increases from 58.0\% to 59.9\%, and the $mAP_{50-95}$ increases from 52.3\% to 53.6\%.
Besides, compared with lines 3 and 6, when we combine the AGF with FEM, the mIoU improves by 0.5\%, and the  $mAP_{50-95}$ improves by 1.6\%.
At the same time, as shown in lines 4 and 8, after adopting the FEM, the mIoU improves by 0.6\%, and the  $mAP_{50-95}$ improves by 0.8\%.
Moreover, when we adopt the FEM, AGF, and TEM, the mIoU improves by 0.7\% and the  $mAP_{50-95}$ improves by 0.4\% compared to adopting AGF and TEM alone, as shown in lines 7 and 9.

In this section, we also explore the influence of different structures within FEM on segmentation and detection tasks.
As described in Section \ref{sec:method}, the feature enhancement module contains the cross-modality spatial integration and the intra-modality channel attention.
We report the experiment results in different settings as shown in Table \ref{tab:FEM}.
In this table, CA represents intra-modality channel attention, SI represents cross-modality spatial integration, and `parallel' and `serial' indicate that we combine CA and SI in parallel and serial configurations, respectively.
As shown in this table, the performance of the semantic segmentation and object detection tasks is optimal when we merge the CA and SI in parallel.

\begin{figure}
    \centering
    \includegraphics[width=0.5\textwidth]{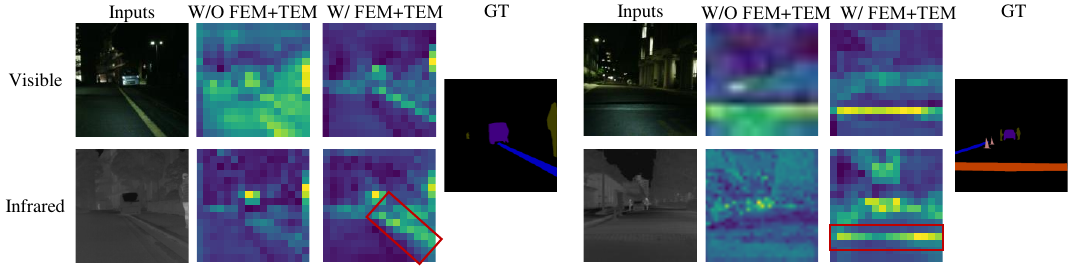}
    \caption{The qualitative analysis of the feature enhancement module.
    `W/O FEM+TEM' indicates that the FEM and TEM modules are not adopted, while `W/ FEM+TEM' indicates that the FEM and TEM modules are included in the proposed method.
    }
    \label{fig:ablation_FEM}
\end{figure}

\begin{table}[]
\centering
\caption{The effectiveness of the feature enhancement module.
`CA' presents the intra-modality channel attention, `SI' presents the cross-modality spatial integration, and `Parallel' and `Serial' mean we combine the `CA' and `SI' through parallel and serial ways.}
\label{tab:FEM}
\resizebox{1\columnwidth}{!}{
\begin{tabular}{cccccc}
\hline
\multicolumn{1}{c}{CA} & \multicolumn{1}{c}{SI} & \multicolumn{1}{c}{Parallel} & \multicolumn{1}{c}{Serial} & \multicolumn{1}{c}{Seg\_mIoU} & \multicolumn{1}{c}{Det\_$mAP_{50-95}$} \\ \hline
\checkmark                        & -                         &     -                         &     -                       &   58.8                           &  53.1                            \\  
-                                 &\checkmark                        & -                      & -                           &  59.2                            &  53.3                            \\  
\checkmark                        & \checkmark                        &  -           &  \checkmark                           &  59.7                             &  53.6                            \\  
\checkmark                        & \checkmark                        &  \checkmark                    & -                 &  59.9                             &   53.8                    \\ \hline
\end{tabular}%
}
\end{table}

\subsubsection{Effect of the Token Enhancement Module (TEM)}
As shown in Table \ref{tab:ablation}, compared with the baseline method, when we only adopt the TEM (line 4) the mIoU increases from 58.0\% to 59.6\%, and the  $mAP_{50-95}$ increases from 52.3\% to 53.6\%.
Additionally, when comparing lines 2 and 8, combining the FEM with TEM results in a 0.3\% improvement in mIoU and a 0.8\% improvement in  $mAP_{50-95}$. 
Furthermore, as shown in lines 3 and 7, adding the TEM leads to a 0.8\% increase in mIoU and a 1.3\% increase in  $mAP_{50-95}$. 
The performance is further enhanced when combining FEM and AGF with TEM, as demonstrated in lines 6 and 9, where the mIoU improves by 1.0\% and the  $mAP_{50-95}$ by 0.1\%. 
These analyses demonstrate the effectiveness of the token enhancement module.

Besides quantitative analyses, we also illustrate some qualitative analysis of FEM and TEM, as can be shown in Fig \ref{fig:ablation_FEM}.
We show the features of $F_{4\_x}$ and $F_{4\_y}$.
`W/O FEM+TEM' indicates that the FEM and TEM modules are not adopted, while `W/ FEM+TEM' indicates that the FEM and TEM modules are included in the proposed method.
We can see that after adding the FEM, the semantic information is more obvious, especially in the infrared modality.
For instance, the `bump' class is barely visible in the infrared images. 
However, the cross-modality spatial integration introduces relevant information from the visible modality into the infrared features.
These figures demonstrate the semantic enhancement capability of the FEM and TEM.

\subsubsection{Effect of Attention-Guided Fusion module (AGF)}
As shown in Table \ref{tab:ablation}, when we only adopt the AGF (line 3) the mIoU increases from 58.0\% to 59.6\%, and the  $mAP_{50-95}$ increases from 52.3\% to 53.8\%.
Besides, compared with lines 2 and 6, when we combine the FEM with AGF, the mIoU improves by 0.2\%, and the  $mAP_{50-95}$ improves by 1.8\%.
At the same time, as shown in lines 4 and 7, after adding the AGF, the mIoU improves by 0.8\%, and the  $mAP_{50-95}$ improves by 1.5\%.
Moreover, when we adopt the FEM, AGF, and TEM, the mIoU improves by 0.9\% and the  $mAP_{50-95}$ improves by 1.1\% compared to adopting FEM and PF alone, as shown in lines 8 and 9.
These analyses demonstrate the effectiveness of the attention-guided fusion module.

Fig. \ref{fig:agf} provides a qualitative analysis of the attention-guided fusion (AGF) module.
The `$F_i$' refers to the output feature of the encoder block-$i$ when the model is trained without AGF, using summing as the fusion method. 
The `$F_i$\_AGF' refers to the output when the AGF module is incorporated into our method.
As we can see from these images, AGF demonstrates a stronger capability to fuse features with richer semantic information.

\begin{figure}
    \centering
    \includegraphics[width=0.5\textwidth]{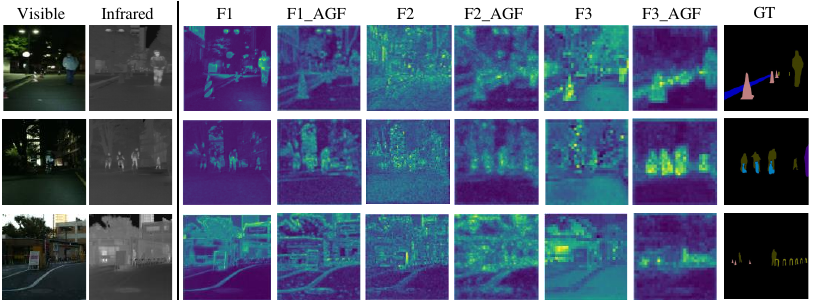}
    \caption{The qualitative analysis of the attention-guided fusion module.
    The F1, F2, and F3 indicate the outputs of the encoder block1, encoder block2, and encoder block3 when we only adopt the summary as the fusion method.
    The F1\_AGF, F2\_AGF, and F3\_AGF are the outputs when we adopt the attention-guided fusion module as the fusion method.
    }
    \label{fig:fusion_agf}
\end{figure}

\subsubsection{Effect of the Cutout\&Mix Augmentation}
As shown in Table \ref{tab:ablation}, when we only adopt the cutout\&mix augmentation (line 5) the mIoU improves by 1.7\%, and the  $mAP_{50-95}$ improves by 1.4\% compared with the baseline method (line 1).
Meanwhile, when we combine the CMA with other modules (lines 9 and 10), the mIoU increases from 61.1\% to 62.4\%, and the  $mAP_{50-95}$ increases from 55.5\% to 55.8\%.
These results show the effectiveness of the cutout\&mix augmentation strategy.


\subsection{Compared with Single-modality Method}
We compare the proposed method with the single-modality methods to demonstrate the effectiveness of fusing two modal images.
We conduct the experiments on semantic segmentation with the MSRS dataset.
Specifically, we use InfMAE \cite{liu2024infmae} as the comparison method for the infrared modality and MCMAE \cite{MCMAE2022} for the visible modality. 
The mIoU of the InfMAE is 74.3\%, the MCMAE is 79.2\%, and the proposed IVGF is 81.0\%.
This results demonstrates that the complementary information of two modal images helps to improve the experimental performance.

\subsection{Discussion on the Situation of Modality missing}
Modal missing is a common challenge in dual-modal tasks.
It occurs when one of the modality images is unavailable due to the environment or imaging equipment.
We report the segmentation performance when either the infrared or visible modality is missing to show the robustness of our proposed method against modal missing.
When one modality is missing, we replace it with the corresponding image from the other modality and conduct these experiments on the MFNet and MSRS datasets.
The experiment results are shown in Table \ref{tab:missing modality}.

From this table, we observe that when the visible modality is missing (line 1), the semantic segmentation performance decreases by 13.2\% and 27.0\%, respectively.
While the infrared modality is missing (line 2), the semantic segmentation performance decreases by 9.8\% and 7.2\%, respectively.
However, compared with the results in Table \ref{tab:MFNet}, Table \ref{tab:MSRS} and Table \ref{tab:missing modality}, we observe that our missing modality results are comparable with other state-of-the-art (SOTA) dual-modality methods.
This demonstrates that our method can handle missing modalities in the segmentation task.
Meanwhile, we observe a significant performance decrease when the visible modality is missing in the MSRS dataset. 
This may be because visible images provide more useful information compared to those in the MFNet dataset.

\begin{table}[]
\centering
\caption{The mIoU is adopted to evaluate the performance of our proposed approach on the MFNet and MSRS datasets under the modality missing situation.}
\label{tab:missing modality}{
\begin{tabular}{ccc}
\hline
Missing Modality   & MFNet  & MSRS\\ \hline
Visible   & 49.2     & 54.0      \\  
Infrared    & 52.6   & 73.8        \\ 
None & 62.4         & 81.0 \\ \hline
\end{tabular}
}
\end{table}

\section{Conclusion} 
\label{sec:conclude}
In this paper, we propose a fusion-guided infrared and visible general framework, named IVGF, which can be easily extended to dual-modal high-level vision tasks.
We adopt our InfMAE and the MCMAE as the foundation models to extract the general representations of infrared and visible modalities.
The feature enhancement and token enhancement modules are designed to enrich the semantic information of general representations.
Additionally, we design an attention-guided fusion module to learn the complementary information from both modalities and fuse them using a cross-attention mechanism.
Besides, we also adopt the cutout\&mix augmentation strategy to conduct the data augmentation, which further improves the ability of the model to mine the regional complementary between the two modalities.
Finally, we validate the effectiveness of the proposed IVGF method in dual-modality semantic segmentation and object detection tasks.
The experimental results show that our method can obtain better experimental results than other SOTA  methods.
In addition, we discuss some ablations in the IVGF to show the effectiveness of each module.
We hope our work will provide some insights into the infrared and visible multimodal and multitasks domains.

\bibliographystyle{ieeetr}
\bibliography{ref}

\end{document}